\documentclass[10pt,twocolumn,letterpaper]{article}

\usepackage[pagenumbers]{cvpr} %

\newcommand{\bI}{\mathbf{I}}

\newcommand{\bs}{\mathbf{s}}

\newcommand{\bv}{\mathbf{v}}
\newcommand{\bq}{\mathbf{q}}

\newcommand{\bp}{\mathbf{p}}

\newcommand{\bC}{\mathbf{C}}

\newcommand{\bff}{\mathbf{f}}
\newcommand{\bF}{\mathbf{F}}

\newcommand{\bh}{\mathbf{h}}

\newcommand{\nR}{\mathbb{R}}

\newcommand{\cD}{\mathcal{D}}

\newcommand{\cS}{\mathcal{S}}

\newcommand{\cV}{\mathcal{V}}
\newcommand{\cY}{\mathcal{Y}}

\definecolor{darkbrown}{rgb}{0.4, 0.26, 0.13}

\newcommand{\figref}[1]{Fig.~\ref{#1}}
\newcommand{\secref}[1]{Sec.~\ref{#1}}

\newcommand{\tabref}[1]{Table~\ref{#1}}

\makeatletter
\DeclareRobustCommand\onedot{\futurelet\@let@token\@onedot}
\def\@onedot{\ifx\@let@token.\else.\null\fi\xspace}
\def\eg{{\em e.g}\onedot}

\def\ie{{\em i.e}\onedot} 
 
 \def\vs{vs\onedot}

\makeatother

\newcommand{\cmark}{\ding{51}}%
\newcommand{\xmark}{\ding{55}}%

\newcommand{\deltaavg}{$\delta_\text{avg}^x$ }

\newcommand{\up}{$\uparrow$}

\definecolor{navyblue}{RGB}{31,119,180}       %
\definecolor{darkorange}{RGB}{255,127,14}     %
\definecolor{forestgreen}{RGB}{44,160,44}     %
\definecolor{crimson}{RGB}{214,39,40}         %
\definecolor{mediumpurple}{RGB}{148,103,189}  %

\definecolor{cvprblue}{rgb}{0.21,0.49,0.74}
\usepackage[pagebackref,breaklinks,colorlinks,allcolors=cvprblue]{hyperref}
\usepackage[dvipsnames, table]{xcolor}
\usepackage{url}            %
\usepackage{booktabs}       %
\usepackage{amsfonts}       %
\usepackage{nicefrac}       %
\usepackage{microtype}      %
\usepackage{url}
\usepackage{pifont}
\usepackage{multirow}
\usepackage{amsmath}
\usepackage{tablefootnote}
\usepackage{tabto}
\usepackage{xspace}        %
\usepackage{graphicx}
\usepackage{subcaption}
\usepackage{xcolor}
\usepackage[toc,page]{appendix} 
\usepackage{cancel}
\usepackage{makecell, cellspace, caption}
\hypersetup{colorlinks,
            linkcolor=NavyBlue,
            urlcolor=NavyBlue,
            citecolor=NavyBlue}
\let\emptyset\varnothing

\def\paperID{} %
\def\confName{CVPR}
\def\confYear{2026}

\title{Real-World Point Tracking with Verifier-Guided Pseudo-Labeling}

\author{Görkay Aydemir\textsuperscript{1} \quad 
        Fatma Güney\textsuperscript{1,2} \textsuperscript{$\dagger$} \quad 
        Weidi Xie\textsuperscript{3} \textsuperscript{$\dagger$} \\[2pt]
        \textsuperscript{1}Department of Computer Engineering, Koç University  \quad
        \textsuperscript{2}KUIS AI Center \\[2pt]
        \textsuperscript{3}School of Artificial Intelligence, 
        Shanghai Jiao Tong University  \\
        \texttt{\small gorkayaydemir@gmail.com}
        }

\begin{document}
\maketitle

\renewcommand{\thefootnote}{\fnsymbol{footnote}}
\footnotetext[2]{ Equal supervision}

\begin{abstract}
Models for long-term point tracking are typically trained on large synthetic datasets. 
The performance of these models degrades in real-world videos due to
different characteristics and the absence of dense ground-truth annotations.
Self-training on unlabeled videos has been explored as a practical solution, but the quality of pseudo-labels strongly depends on the reliability of teacher models, which vary across frames and scenes.
In this paper, we address the problem of real-world fine-tuning and introduce \textbf{verifier}, a meta-model that learns to assess the reliability of tracker predictions and guide pseudo-label generation. Given candidate trajectories from multiple pretrained trackers, the verifier evaluates them per frame and selects the most trustworthy predictions, resulting in high-quality pseudo-label trajectories. When applied for fine-tuning, verifier-guided pseudo-labeling substantially improves the quality of supervision and enables data-efficient adaptation to unlabeled videos.
Extensive experiments on four real-world benchmarks demonstrate that our approach achieves state-of-the-art results while requiring less data than prior self-training methods. Project page: \href{https://kuis-ai.github.io/track_on_r}{\texttt{kuis-ai.github.io/track\_on\_r}}.
\end{abstract}

\section{Introduction}
\label{sec:intro}

\begin{figure*}[t]
    \centering
    \includegraphics[width=1\linewidth]{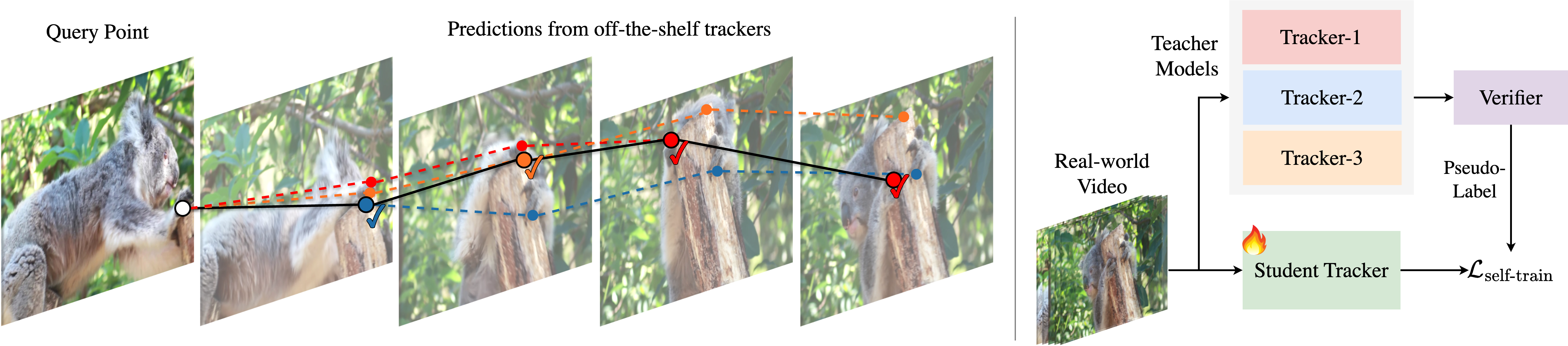}
    \caption{\textbf{Verifier-guided real-world adaptation.} 
    \textbf{(Left)} Given a query point in a real-world video, multiple off-the-shelf trackers produce alternative trajectory hypotheses. \textbf{Verifier} evaluates these per-frame predictions and selects the most reliable ones, forming a refined pseudo-label trajectory. 
    \textbf{(Right)} Unlike naïve self-training, which randomly selects a teacher model for pseudo-label generation, the verifier adaptively combines predictions from multiple teachers, providing cleaner supervision for the student tracker during real-world fine-tuning.
    }
    \label{fig:verifier_teaser}
\end{figure*}

Motion estimation is a longstanding problem in computer vision, aiming to reliably track physical points across video frames. While short-term correspondences can be well handled by optical flow~\cite{Sun2018CVPR, Teed2020ECCV}, extending this capability over long temporal horizons, often referred to as point tracking~\cite{Harley2022ECCV, Doersch2022NeurIPS}, has recently drawn increased attention. Point tracking is a fundamental primitive for long-term visual understanding and control, enabling a wide range of applications where motion is crucial, including video editing~\cite{Geng2025CVPR}, robot manipulation and perception~\cite{Bharadhwaj2024ECCV, Wen2023ARXIV, Vecerik2023ICRA}, 4D scene understanding~\cite{Rockwell2025CVPR, Badki2025ARXIV, Wang2025CVPR, Kasten2024NeurIPS, Feng2025ICCV}, and medical analysis~\cite{Azad2025ICCVW, Karaoglu2025MICCAI}.

Recently, transformer-based point trackers~\cite{Karaev2024ECCV, Li2024ECCV, Aydemir2025ICLR} have advanced the state of the art,
yet their training regimes remain synthetic due to the prohibitive cost of dense, frame-accurate long-term trajectory annotation in real videos. As a result, models often inherit a sim-to-real discrepancy: appearance statistics, nonrigid motion, occlusion patterns, lighting changes, and sensor artifacts in natural footage degrade reliability over extended sequences. Bridging this gap requires training that can exploit unlabeled, large-scale real-world videos without access to dense ground-truth trajectories.

Self-training via pseudo-labels is an attractive path forward: predictions from a pretrained tracker (or an ensemble) serve as supervision to adapt on real data~\cite{Karaev2024ARXIV}. However, naïve pseudo-labeling is brittle, as the teacher predictions are not uniformly reliable. Different trackers excel in different regimes, some resist fast motion but drift under low texture; others handle occlusion better but suffer from identity switches or jitter. Fixed heuristics or global confidence thresholds cannot reconcile these heterogeneous error profiles and often propagate systematic errors during adaptation.

We argue that effective real-world training for point tracking hinges on reliability estimation: models must learn when and where to trust tracker outputs. To this end, we introduce a \textbf{verifier}, a learned meta-model that scores the framewise reliability of candidate trajectories produced by multiple pretrained, off-the-shelf trackers. Given a query point and several candidate tracks, the {verifier} predicts which candidate most faithfully follows the underlying motion at each frame, enabling dynamic selection and seamless switching as conditions change~(\figref{fig:verifier_teaser}, left). Trained entirely on synthetic data with ground-truth trajectories, the {verifier} learns from deliberately perturbed candidates that emulate realistic errors (drift, jumps, occlusions, re-appearances), using a contrastive objective to rank correct versus corrupted alternatives. Crucially, this training requires no real-world annotations yet teaches the {verifier} to recognize consistency cues that transfer across domains.

During adaptation on real videos, the verifier acts as a supervision selector: it filters and fuses per-frame reliability scores to form robust pseudo-label trajectories, reducing error accumulation and preventing collapse when any single teacher fails~(Fig.~\ref{fig:verifier_teaser}, right). The same mechanism can be used at inference time as a plug-and-play ensemble module, combining complementary trackers on the fly based on learned reliability cues rather than fixed weighting. This unified approach turns model diversity into a strength, yielding annotation-free adaptation, improved robustness to distribution shift, and better long-term coherence.

We validate our approach across diverse real-world datasets and tracking regimes. Verifier-guided pseudo-labeling consistently improves effectiveness during fine-tuning on unlabeled videos, and verifier-based ensembling at inference provides additional gains, particularly under challenging motion and occlusion. Ablations confirm that the verifier effectively exploits the complementary strengths of different trackers while remaining robust to individual failures. Together, these results show that the verifier provides a unified, data-efficient framework for reliable pseudo-labeling and model coordination in real-world point tracking.

In summary, our contributions are threefold:
(i) a verifier, that learns to score and select reliable per-frame predictions from multiple trackers, enabling both training-time supervision selection and optional inference-time ensembling; 
(ii) a verifier-guided pseudo-labeling framework that scales fine-tuning on real videos while mitigating the failure modes of naïve pseudo-labeling;
(iii) extensive experiments and ablations showing state-of-the-art performance and robust gains across real-world point-tracking benchmarks, highlighting the data efficiency and practicality of our approach.
\section{Related Work}
\label{sec:rw}

 \begin{figure*}[t]
  \centering
  \begin{subfigure}[t]{0.265\linewidth}
    \centering
    \includegraphics[width=\linewidth,trim=4 6 4 0,clip]{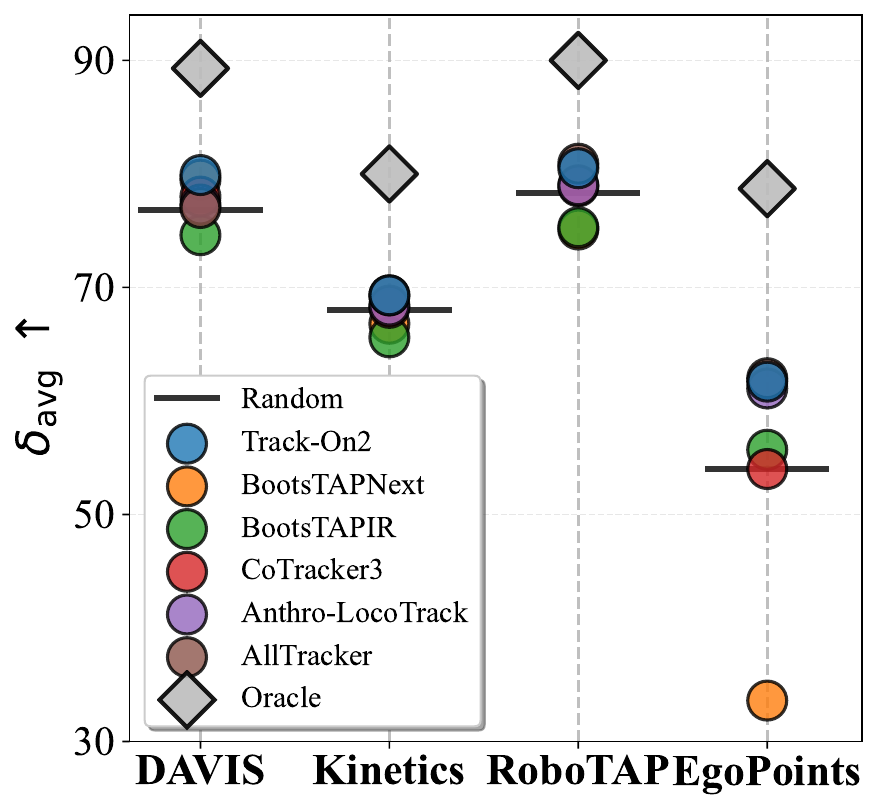}
    \caption{\textbf{Oracle comparison across datasets.}}
    \label{fig:teacher-oracle-a}
  \end{subfigure}
  \hfill
  \begin{subfigure}[t]{0.73\linewidth}
    \centering
    \includegraphics[width=\linewidth,trim=0 6 0 2,clip]{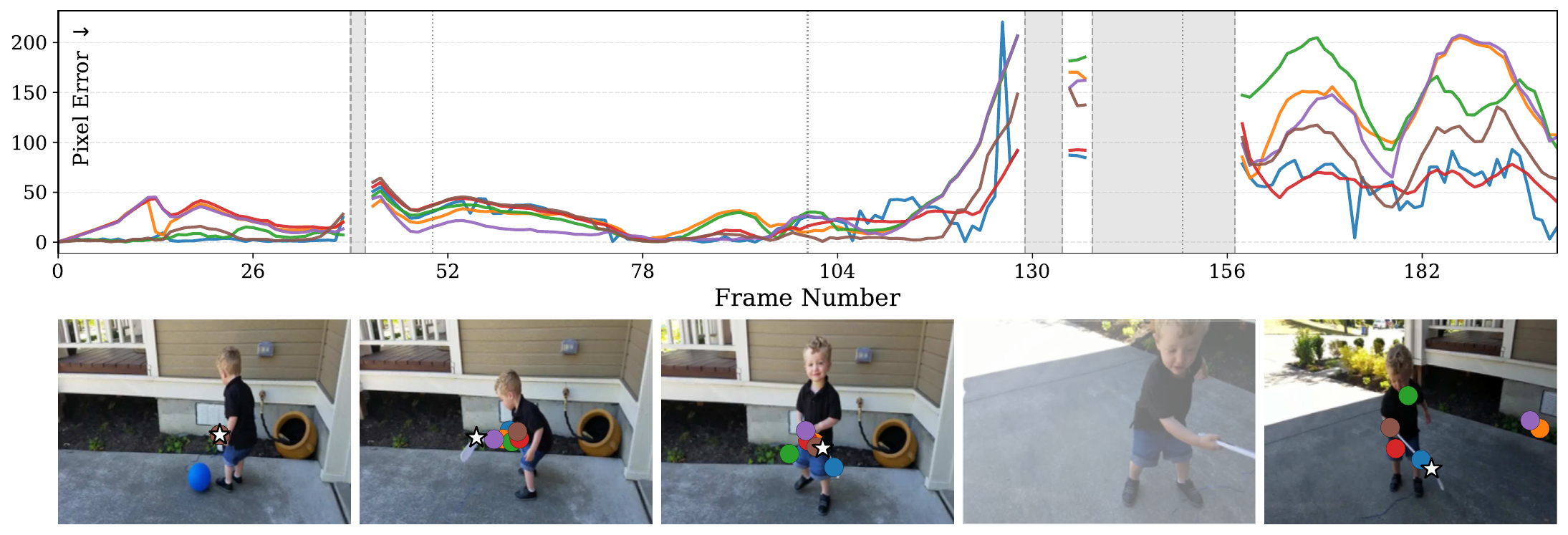}
    \caption{\textbf{Teacher disagreement across frames.}}
    \label{fig:teacher-oracle-b}
  \end{subfigure}
    \caption{\textbf{Teacher inconsistency and oracle performance.}
    \textbf{(a)}~Across 4 real-world datasets, six off-the-shelf teacher models (shown on the legend) are compared against an oracle that, at each frame, selects the most accurate teacher prediction. 
    Individual teachers (colored circles) cluster below the oracle (diamonds), while the black horizontal line marks the performance of random teacher selection. 
    The large gap between the oracle and both individual models and random selection highlights the substantial headroom available for adaptive, per-frame selection.
    \textbf{(b)}~Example from TAP-Vid Kinetics~\cite{Doersch2022NeurIPS}: Teacher predictions whose pixel errors fluctuate across time. 
    The upper plot shows per-frame pixel error curves, with occluded frames shaded in gray. 
    Colored lines correspond to the same trackers as in (a), illustrating that accuracy varies across time.
    The lower row shows uniformly sampled frames with teacher predictions and the ground-truth point (white star).}
  \label{fig:teacher-oracle}
\end{figure*}

\noindent\textbf{Point tracking.}
Tracking arbitrary points across long videos requires maintaining fine-grained correspondences under motion, occlusion, and reappearance. PIPs~\cite{Harley2022ECCV} introduced iterative refinement within temporal windows, and TAP-Vid~\cite{Doersch2022NeurIPS} established a large-scale benchmark. TAPIR~\cite{Doersch2023ICCV} improved temporal precision, while CoTracker~\cite{Karaev2024ECCV} reformulated tracking as joint multi-point reasoning with transformers. Subsequent works extended these ideas through query-based~\cite{Li2024ECCV, Li2024NeurIPS} or region-level designs~\cite{Cho2024ECCV}, later unified in CoTracker3~\cite{Karaev2024ARXIV}. Track-On~\cite{Aydemir2025ICLR} recast tracking as patch classification for online inference, and Track-On2~\cite{Aydemir2025ARXIV} enhanced its efficiency. TAPNext~\cite{Zholus2025ICCV} explored a complementary state-space approach. In contrast, we focus on improving real-world reliability through ensemble-based adaptation rather than new architecture design.

\vspace{3pt}
\noindent\textbf{Pseudo-labeling for unlabeled data.}
Pseudo-labeling exploits model predictions as supervision for unlabeled data~\cite{Lee2013ICMLWORK, Tarvainen2017NeurIPS, Xie2020CVPR, Sohn2020NeurIPS}, while ensemble learning~\cite{Freund1996ICML, Breiman1996ML} combines predictors to improve generalization. Recent variants~\cite{Lakshminarayanan2017NeurIPS, Gustafsson2020CVPRW, Lu2024ICML, Chen2023NeurIPS} show that structured consensus across models yields more reliable supervision than single-teacher schemes. In point tracking, pseudo-labels bridge the gap between synthetic and real domains: BootsTAPIR~\cite{Doersch2024ARXIV} applies large-scale self-distillation, CoTracker3~\cite{Karaev2024ARXIV} integrates real pseudo-labeled data, and AnthroTAP~\cite{Kim2025ARXIV} uses 3D mesh priors for human motion. Our verifier extends this paradigm into a learnable ensemble, reasoning across trackers to select reliable labels via spatio-temporal cues.
\section{Problem Formulation}
\label{sec:method}

Given an RGB video of $T$ frames, 
$\cV = \{\bI_1, \dots, \bI_T\} \in \nR^{T \times H \times W \times 3}$, 
and a query point $\bq_{t_0} \in \nR^2$ at time $t_0$, 
the goal of point tracking is to predict its trajectory and visibility across subsequent frames:
\begin{equation}
\{ ( \hat{\bp}_t, \hat{v}_t ) \}_{t=t_0+1}^T = \Phi(\cV, \bq_{t_0})
\end{equation}
where $\hat{\bp}_t \in \nR^2$ denotes the predicted 2D coordinates 
and $\hat{v}_t \in \{0,1\}$ indicates whether the point is visible at time $t$.

\vspace{3pt} \noindent \textbf{Data setup \& training regime.}
From the data perspective, we consider two domains: $\cD_{\text{syn}}$ and $\cD_{\text{real}}$. 
The labeled synthetic dataset $\cD_{\text{syn}} = \{(\cV, \cY)\}$ contains videos and their ground-truth point trajectories, 
where $\cY = \{(\bp_t, v_t)\}_{t=1}^{T}$ denotes the trajectory and visibility of one queried point. 
In practice, multiple query points can be sampled per video, 
each with its own trajectory. 
The unlabeled real-world dataset $\cD_{\text{real}} = \{\cV\}$ consists of videos without trajectory annotations, \ie $\cY = \emptyset$.

Existing approaches fall into two regimes:  
(i) training directly on $\cD_{\text{syn}}$; and  
(ii) pretraining on $\cD_{\text{syn}}$ followed by self-training on $\cD_{\text{real}}$ using pseudo-labels. 

\vspace{3pt} \noindent \textbf{Sim2Real training on $\cD_{\text{syn}}$.}
Most existing point tracking models are trained on large-scale synthetic datasets $\cD_{\text{syn}}$, such as TAP-Vid Kubric~\cite{Doersch2022NeurIPS}, and evaluated on both synthetic~\cite{Karaev2023CVPR, Zheng2023ICCV} and real-world benchmarks~\cite{Vecerik2023ICRA, Doersch2022NeurIPS}. 
This setting assumes that models trained on synthetic videos can generalize to real ones, yet in practice, differences in texture, illumination, motion, and occlusion patterns introduce a domain gap, potentially leading to degraded performance.

\vspace{3pt} \noindent \textbf{Naïve self-training on $\cD_{\text{real}}$.} To bridge the domain gap, existing work performs self-training on real videos using pseudo-labels generated by pretrained teacher models~\cite{Karaev2024ARXIV}. Formally, a video $\cV \sim \cD_{\text{real}}$ is sampled, and a teacher $\Phi^{(m)}$ is randomly selected from a set of $M$ pretrained models $\{\Phi^{(1)}, \dots, \Phi^{(M)}\}$. For randomly chosen query point $\bq_{t_0}$, pseudo-labels $\tilde{\cY} = \{(\hat{\bp}_t, \hat{v}_t)\}_{t=1}^{T} = \Phi^{(m)}(\cV, \bq_{t_0})$ are then used as supervision in place of the ground-truth $\cY$.

\vspace{3pt} \noindent \textbf{Discussion.}
A single teacher rarely provides uniformly reliable predictions across an entire video; accuracy fluctuates frame to frame, with intermittent successes and failures (Fig.~\ref{fig:teacher-oracle-b}). In practice, different trackers exhibit error patterns that are inconsistent and only weakly correlated, yielding temporally unstable pseudo-labels rather than truly complementary signals. Naively sampling from, or averaging across, such predictions tends to amplify noise and drift, undermining the benefits of self-training on real videos (Fig.~\ref{fig:teacher-oracle-a}).
We address these limitations with a verifier that explicitly estimates per-frame reliability and selects among candidate trajectories accordingly. By conditioning supervision on learned reliability rather than fixed heuristics, the verifier stabilizes pseudo-labels, suppresses failure cascades, and turns model diversity into a practical advantage for adaptation.

\begin{figure*}[t]
    \centering
    \includegraphics[width=1\linewidth]{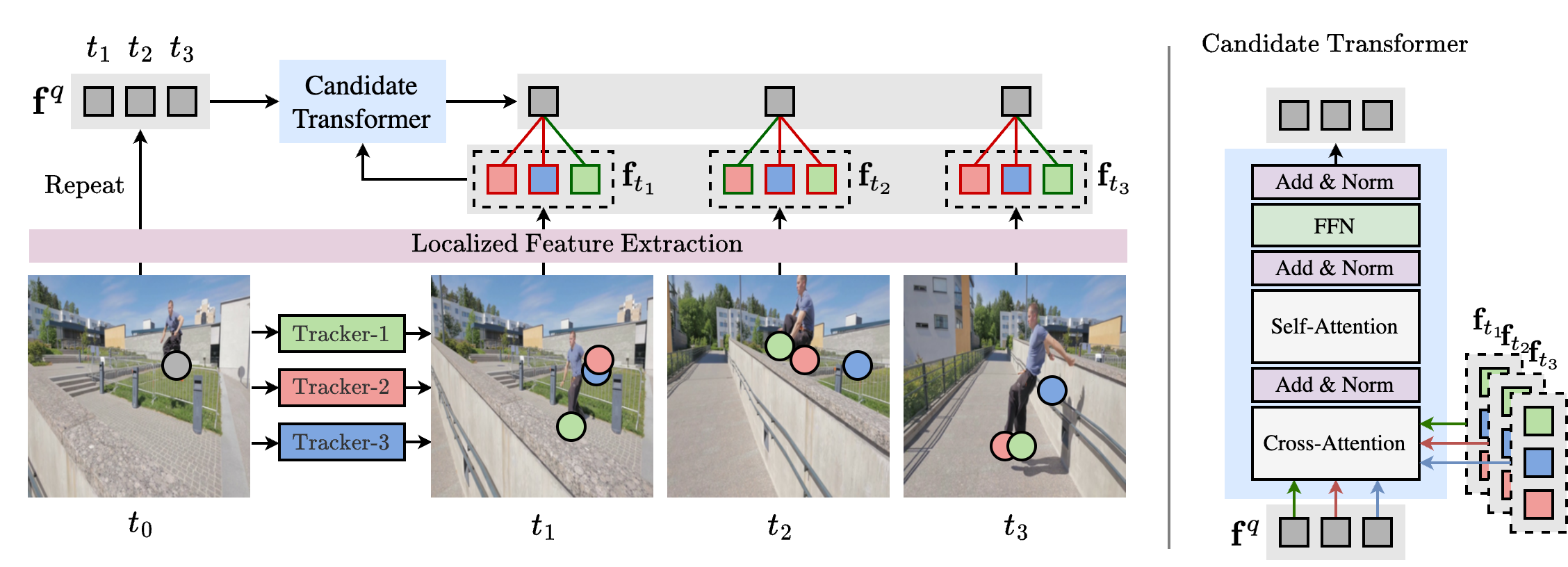}
    \caption{\textbf{Verifier overview.} 
    Given query points at frame $t_0$ and their candidate predictions (teacher outputs during inference or randomly augmented trajectories during training), we extract local features for both the query and each candidate, producing query features $\bff^q$ (replicated across time) and candidate features $\bff_t$ (a vector for each candidate, per frame). The query features are then decoded by the Candidate Transformer (\textbf{right}), which consists of restricted cross-attention, where each frame-level query attends only to its corresponding candidates, followed by self-attention and feed-forward layers. The transformer outputs per-frame reliability distributions over candidates, capturing spatial and temporal consistency based on feature similarity.}
    \label{fig:verifier_architecture}
\end{figure*}

\section{Verifier}
\label{sec:verifier}

We introduce a \textbf{verifier}, trained on labeled synthetic videos, to assess the reliability of tracker predictions. 
At inference, it operates on real-world videos to select, at each time step, the most reliable estimate among multiple pretrained teacher trackers. Unlike conventional trackers, the verifier is a meta-model,
{\em i.e.}, it does not predict trajectories directly but learns to evaluate which tracker is most reliable for each frame. 
Specifically, it compares the visual features around the predicted locations with those of the initial query point, reasoning across teachers and time to estimate a per-frame reliability score for each candidate trajectory. These scores are then used to construct a refined trajectory by combining the most reliable predictions over time, producing stable and accurate pseudo-labels for real-world fine-tuning.

\vspace{3pt}
\noindent \textbf{An oracle test on existing tracking models.}
To measure the available headroom when combining multiple teachers, we construct an \textit{oracle} that, at each time step, selects the prediction closest to the ground truth (\figref{fig:teacher-oracle-a}).
As expected, the oracle performs best among all settings, but the large gap between it and the individual teachers highlights a significant potential for improvement. Across datasets, this gap remains consistent and is particularly pronounced on challenging egocentric videos such as EgoPoints~\cite{Darkhalil2025WACV}. These results indicate that the best-performing teacher varies across frames and scenes, making static or random selection suboptimal. Motivated by this observation, we propose to train a verifier that adaptively identifies the most reliable prediction among off-the-shelf trackers, producing more stable and accurate pseudo-labels $\tilde{\mathcal{Y}}$ on $\cD_{\text{real}}$.

Formally, the verifier takes as input a query point $\bq_{t_0}$ in a video $\cV$ and a tensor of $M$ \textbf{candidate trajectories} that provide alternative motion estimates for the same target.  
We represent these trajectories as $\bC \in \nR^{L \times M \times 2}$,  
where $L = T - t_0$ is the trajectory length and each $\hat{\bp}_t^{(m)}$ specifies the 2D location of candidate $m$ at frame $t$.  
The verifier assigns a reliability score to each candidate at every time step, producing a set of per-frame score vectors:
\begin{equation}
\hat{\cS} = \Phi_{\text{ver}}(\cV, \bq_{t_0}, \bC)
\end{equation}
where $\hat{\cS} = \{\hat{\bs}_t \mid t \in (t_0, T]\}$,  
each $\hat{\bs}_t \in \nR^{M}$ contains the reliability scores of all candidates at frame $t$,  
and $\hat{s}_t^{(m)}$ denotes the score assigned to candidate $m$.  
We describe how $\bC$ is constructed in~\secref{sec:verifier:candidates}.  
As illustrated in~\figref{fig:verifier_architecture}~(left), the verifier first extracts localized features for the query and all candidate trajectories at each frame (\secref{sec:verifier:local_encoder}).  
The candidate transformer then reasons jointly over candidates and time to identify the most consistent motion hypothesis (\secref{sec:verifier:transformer}).  
Finally, the verifier outputs per-frame reliability scores, selecting the most plausible candidate at each frame.  
These selected predictions are subsequently used to fine-tune a tracking model for real-world adaptation (\secref{sec:verifier:fine_tuning}).

\subsection{Candidate Trajectories}
\label{sec:verifier:candidates}

\noindent \textbf{Training setup.}  
During training, the verifier is supervised on labeled synthetic videos $(\cV, \cY) \sim \cD_{\text{syn}}$.
Each ground-truth trajectory is represented by $\bp \in \nR^{L \times 2}$, where $L$ is the trajectory length and $\bp_t$ gives the 2D location of the tracked point at frame $t$.
The candidate trajectories $\bC$ are generated by applying random perturbations to $\bp$, producing $M$ (typically 6–12) perturbed versions, \ie $\bp \in \nR^{L \times 2} \rightarrow \bC \in \nR^{L \times M \times 2}$, that simulate common prediction errors such as drift, occlusion, and jitter (see supplementary for details).
These augmentations create diverse motion hypotheses resembling the errors observed at inference, allowing the verifier to learn to distinguish reliable trajectories from unreliable ones.
Each candidate’s reliability target is computed from its distance to the corresponding ground-truth point, providing explicit supervision for training the verifier to rank predictions by reliability, as detailed in~\secref{sec:verifier:training}.

\vspace{3pt}
\noindent\textbf{Inference setup.} 
At inference and during real-world fine-tuning, $\bC$ consists of trajectories predicted by $M{=}6$ pretrained teacher trackers: Track-On2~\cite{Aydemir2025ARXIV}, BootsTAPIR~\cite{Doersch2024ARXIV}, BootsTAPNext~\cite{Zholus2025ICCV}, Anthro-LocoTrack~\cite{Kim2025ARXIV}, AllTracker~\cite{Harley2025ICCV}, and CoTracker3 (window variant)~\cite{Karaev2024ARXIV}.  
For each query point, the verifier evaluates these trajectories and assigns frame-wise reliability scores, selecting the highest-confidence prediction as the pseudo-label used for fine-tuning on $\cD_{\text{real}}$ without manual annotations (\secref{sec:verifier:fine_tuning}).

\subsection{Localized Feature Extraction}
\label{sec:verifier:local_encoder}

The verifier determines, at each frame, which candidate trajectory best continues the motion of the query point. To make this judgment, it compares the local visual evidence around the query with the appearance context around each candidate prediction. The localized feature extraction therefore converts image regions near these 2D locations into compact descriptors that are directly comparable.

\vspace{3pt}
\noindent \textbf{Frame-wise feature extraction.}
We compute dense visual features for all frames using the frozen CNN encoder of pretrained CoTracker3~\cite{Karaev2024ARXIV}, 
projected to dimension $D$ through a linear layer:
\begin{equation}
\bF_t = \phi_\text{enc}(\bI_t), \quad \bF_t \in \mathbb{R}^{H' \times W' \times D},
\end{equation}
where $(H', W')$ are the downsampled spatial resolutions.

\vspace{3pt}
\noindent \textbf{Query and candidate representations.}
The verifier compares each candidate prediction against a stable visual reference of the target.
The query point serves as this reference, encoding the appearance and context of the target at its first visible frame $t_0$.
We first obtain this reference embedding by bilinearly sampling the feature map at $\bq_{t_0}$:
\begin{equation} \label{eq:feature_sampling}
\bq_{\text{sample}} = \text{sample}(\bF_{t_0},~\bq_{t_0}) \in \mathbb{R}^{D}
\end{equation}
For each frame $t$, the verifier measures how well the local region around each candidate in $\bC_t \in \nR^{M \times 2}$ matches the query reference.  
To capture this local context, deformable attention is applied at both the query and candidate locations, producing locally aggregated descriptors:
\begin{align} \label{eq:def_attn}
\bh_{t_0}^q &= 
\phi_{\text{def}}(\bq_{\text{sample}}, \bF_{t_0}, \bq_{t_0}) \in \nR^{D}, \\
\bh_t &= 
\phi_{\text{def}}(\bq_{\text{sample}}, \bF_t, \bC_t) \in \nR^{M \times D}, 
\quad t > t_0.
\end{align}
Here, $\phi_{\text{def}}(\cdot)$ uses the reference feature $\bq_{\text{sample}}$ as the query input, the frame feature map as key–value pairs, and the spatial coordinates from $\bC_t$ as attention centers, ensuring that all candidate descriptors are computed relative to the same query appearance.  
In essence, these features capture how the appearance of the initial query is expressed across all candidate locations.  
This operation aggregates adaptive contextual information around each candidate, enabling the verifier to compare appearance similarity within local neighborhoods rather than relying on single-point features.

\vspace{3pt}
\noindent\textbf{Position and identity embeddings.}  
The extracted features are purely visual and lack explicit spatial context.
To provide positional context, we apply a sinusoidal embedding $\eta(\cdot)$ to displacement vectors,  
and append a learned identity embedding that distinguishes the query ($\text{ID}_0$) from the candidate predictions ($\text{ID}_1$).  
After concatenation, a projection layer $\phi_{\text{proj}}$ maps the combined representation to the model width:
\begin{align} \label{eq:local_feature}
\bff^q_{t_0} &= \phi_{\text{proj}}\bigl(\bh^q_{t_0},~\eta(\mathbf{0}),~\text{ID}_0\bigr) \in \nR^{D}, \\
\bff_t &= \phi_{\text{proj}}\bigl(\bh_t,~\eta(\boldsymbol{\Delta}_t),~\text{ID}_1\bigr) \in \nR^{M \times D}, 
\quad t > t_0,
\end{align}
where $\boldsymbol{\Delta}_t = \bC_t - \bq_{t_0}$ represents the displacement of each candidate from the initial query location,  
and $\phi_{\text{proj}}$ aligns the concatenated representation to the model dimension.

\vspace{3pt}
\noindent\textbf{Output.}  
The local encoder produces temporally aligned feature descriptors for both the query and candidate trajectories.  
The query feature $\bff_{t_0}^q$ is replicated across time,  
\ie $\bff_{t_0}^q \!\rightarrow\! \{\bff_{t_0+1}^q, \dots, \bff_{T}^q\}$,  
forming a tensor $\bff^{q} \in \nR^{L \times D}$ that encodes the reference appearance of the target at each frame,  
where each $\bff_t^{q} \in \nR^{D}$ corresponds to the query embedding at frame $t$.  
The candidate descriptors across all frames are stacked into a tensor $\bff \in \nR^{L \times M \times D}$,  
with $\bff_t \in \nR^{M \times D}$ representing the $M$ candidate features at frame $t$.
Together, $\bff^{q}$ and $\bff$ form the input to the candidate transformer.

\subsection{Candidate Transformer}
\label{sec:verifier:transformer}

The candidate transformer decodes the query descriptors $\bff^{q}$ with the candidate descriptors $\bff$ to produce temporally informed embeddings for reliability estimation.
It extends a standard transformer decoder~\cite{Vaswani2017NeurIPS} to reason jointly over candidates and time.  
Each layer consists of localized cross-attention, temporal self-attention, and a feed-forward network (\figref{fig:verifier_architecture}, right).

At each frame $t$, the query embedding $\bff_t^{q} \in \nR^{D}$ attends only to its corresponding candidate features $\bff_t \in \nR^{M \times D}$,
performing cross-attention along the candidate dimension $M$.
During this step, the temporal dimension $L$ is treated as a batch axis,
so attention is computed independently for each frame.
This step allows the query to integrate information from all candidate trajectories at that frame and identify the most consistent motion hypothesis.
The resulting per-frame query embeddings are then connected across time through self-attention along the temporal dimension $L$,
allowing information flow between frames.
This temporal reasoning maintains consistency and helps resolve ambiguities:
if a prediction is uncertain at one frame, contextual evidence from nearby frames with confident predictions can refine its reliability estimate.
The outputs of these attention stages are passed through a feed-forward network to produce updated query representations for the next layer.
No attention mask is applied, and all candidate predictions are included even when the point is occluded at that frame.

\vspace{3pt}
\noindent\textbf{Verifier output.}  
After decoding the query embeddings, the candidate transformer outputs a reliability distribution over candidates at each time step.  
Formally, for each frame $t$, we compute a temperature-scaled softmax over cosine similarities between the decoded query feature $\bff_t^{q}$ and the candidate features $\bff_t$:
\begin{equation}
\hat{\bs}_t =
\mathrm{Softmax}\!\left(\bff_t^{q} \cdot \bff_t / \tau \right),
\qquad \hat{\bs}_t \in \nR^{M}.
\end{equation}
Here, $\bff_t \in \nR^{M \times D}$ represents all candidate features at frame $t$,  
$\bff_t^{q} \in \nR^{D}$ is the decoded query feature,  
and $\cdot$ denotes cosine similarity applied row-wise between $\bff_t^{q}$ and each row of $\bff_t$, scaled by the temperature $\tau = 0.1$.  
This yields a per-frame reliability distribution $\hat{\bs}_t$ over the $M$ candidates.

\begin{figure*}[t]
    \centering
    \includegraphics[width=1\linewidth]{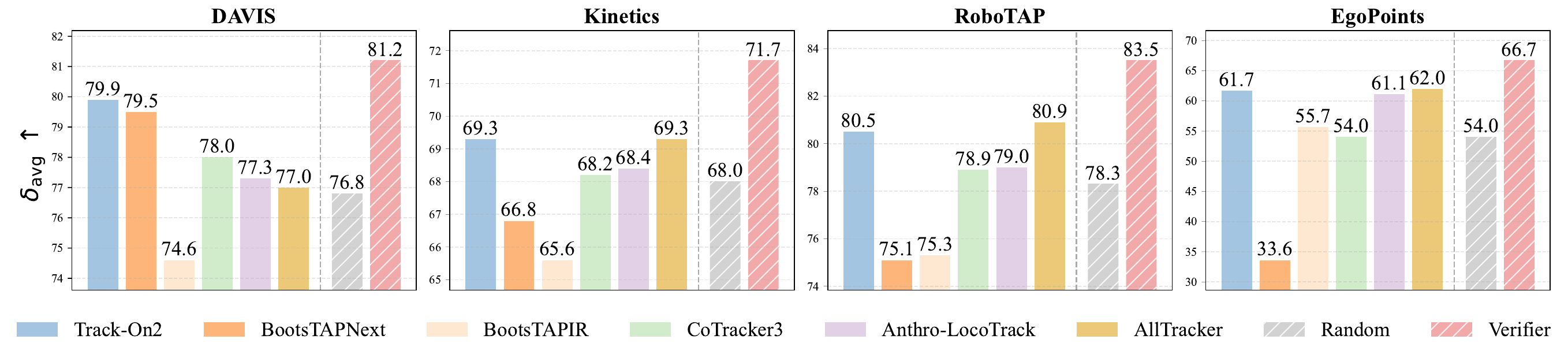}
    \caption{\textbf{Verifier as inference time ensemble.} Comparison of the verifier ensemble against individual teacher models and the random-teacher baseline on real-world datasets. All teacher results are reproduced using their official checkpoints. The verifier consistently achieves the best performance across datasets, demonstrating its ability to exploit the complementary strengths of different models.}
    \label{fig:verifier_comparison}
\end{figure*}

\subsection{Training}
\label{sec:verifier:training}

Given the ground-truth trajectory $\bp \in \nR^{L \times 2}$ and its visibility sequence $\bv \in \{0,1\}^{L}$,  
the verifier is trained with a soft contrastive objective.  
Intuitively, candidate predictions that lie closer to the ground-truth locations should receive higher reliability,  
while those farther away should be downweighted.  
At each frame $t$, the verifier receives a set of $M$ candidate predictions $\bC_t \in \nR^{M \times 2}$  
and the corresponding ground-truth point $\bp_t \in \nR^{2}$.  
We define the per-frame target reliability distribution $\bs_t$  
as a softmax over the negative distances between all candidates in $\bC_t$ and $\bp_t$:
\begin{equation}
\bs_t = \mathrm{Softmax}\!\left(-\lVert \bC_t - \bp_t \rVert / \tau_s \right),
\qquad \bs_t \in \nR^{M},
\end{equation}
where $\lVert \bC_t - \bp_t \rVert$ computes the Euclidean distance from each candidate to the ground truth,  
and $\tau_s = 0.3$ controls the sharpness of the distribution.  
The predicted reliability scores $\hat{\bs}_t$ are supervised using a cross-entropy loss  
$\mathcal{L} = \sum_{t=1}^{L} v_t \, \mathrm{CE}(\hat{\bs}_t, \bs_t)$,  
where $v_t$ masks out occluded frames, ensuring that only visible points contribute to the objective.

\subsection{Fine-tuning with Verified Pseudo-Labels}
\label{sec:verifier:fine_tuning}

We start from a point tracking model pretrained on labeled synthetic data $\cD_{\text{syn}}$ and adapt it to real-world videos from $\cD_{\text{real}}$.  
In our case, the base model is Track-On2~\cite{Aydemir2025ARXIV}, pretrained on TAP-Vid Kubric~\cite{Doersch2022NeurIPS}. 
For real-world adaptation, $\cD_{\text{real}}$ is constructed from in-the-wild object tracking and segmentation datasets, namely TAO~\cite{Dave2020ECCV}, OVIS~\cite{Qi2022IJCV}, and VSPW~\cite{Miao2021CVPR}. 
We retain videos longer than 48 frames without additional filtering, resulting in 4864 real-world sequences with diverse motion patterns and scene content.

To fine-tune the model on $\cD_{\text{real}}$, we employ the trained verifier to generate reliable pseudo-labels. 
For each video $\cV \sim \cD_{\text{real}}$, we sample query points based on both appearance and motion cues. 
Specifically, 2/3 of the queries are drawn from SIFT~\cite{Lowe1999ICCV} detections, while the remaining 1/3 are selected from motion-salient regions obtained via grayscale frame differencing with mild spatial smoothing. 
Queries are sampled from the first half of the video, using four uniformly spaced frames (\eg, frames 1, 5, ...). 

Each query is tracked by the ensemble of teacher models, producing a set of candidate trajectories. 
At every frame $t$, the candidate with the highest reliability score in $\hat{\bs}_t$, \ie the prediction corresponding to $\arg\max_\text{m} \hat{s}_t^{(m)}$, is selected as the pseudo-label. 
Visibility is estimated via majority voting across teacher predictions. This verifier-guided pseudo-labeling enables efficient adaptation to real-world domains without requiring manual annotations or labeled data.

In practice, we fine-tune using both $\cD_{\text{syn}}$ (with ground-truth annotations) and $\cD_{\text{real}}$, gradually increasing the loss weight of real-world samples while reducing that of synthetic ones during training. 
This schedule yields slightly better performance than training on $\cD_{\text{real}}$ alone, although using only $\cD_{\text{real}}$ still produces competitive results (see~\secref{sec:exp:ablation}).

\section{Experiments}
\label{sec:exp}

\begin{table*}
    \centering
    \caption{\textbf{Quantitative results on real-world datasets.}
    Comparison with prior work on EgoPoints, RoboTAP, TAP-Vid Kinetics, and TAP-Vid DAVIS in terms of AJ, \deltaavg, and OA (higher is better).
    Models are grouped into \textbf{synthetic pretraining} and \textbf{real-world fine-tuning}.
    The latter includes methods fine-tuned on additional real-world videos using semi-supervised strategies. We report officially published results when available; if no reported values exist, we evaluate the authors’ official checkpoints. Missing checkpoints are shown as “—”. \\ $^\dagger$ AllTracker leverages additional optical-flow datasets (synthetic and real-world with ground-truth), without pseudo-label fine-tuning.
    }
    \label{tab:sota_A}

    \begin{tabular}{l cc ccc ccc ccc}
        \toprule
        \multirow{3}{*}{\textbf{Model}} 
        & \multicolumn{2}{c}{EgoPoints} 
        & \multicolumn{3}{c}{RoboTAP} 
        & \multicolumn{3}{c}{Kinetics} 
        & \multicolumn{3}{c}{DAVIS} \\
        \cmidrule(r){2-3} \cmidrule(r){4-6} \cmidrule(r){7-9} \cmidrule(r){10-12}
         & \deltaavg & OA 
         & AJ & \deltaavg & OA 
         & AJ & \deltaavg & OA 
         & AJ & \deltaavg & OA \\
        \midrule

        \textbf{Synthetic Pretraining} \\
        TAPIR~\cite{Doersch2023ICCV} & 50.2 & 79.9 & 59.6 & 73.4 & 87.0 & 49.6 & 64.2 & 85.0 & 56.2 & 70.0 & 86.5 \\
        LocoTrack~\cite{Cho2024ECCV} & 59.6 & 88.6 & 62.3 & 76.2 & 87.1 & 52.9 & 66.8 & 85.3 & 63.0 & 75.3 & 87.2 \\ 
        TAPNext-B~\cite{Zholus2025ICCV} & 31.8 & 73.8 & 59.5 & 72.8 & 88.0 & 53.3 & 67.9 & 87.0 & 62.4 & 76.6 & 90.5 \\
        CoTracker3 (Window)~\cite{Karaev2024ARXIV} & --- & --- & 60.8 & 73.7 & 87.1 & 54.1 & 66.6 & 87.1 & 64.5 & 76.7 & 89.7 \\
        Track-On2~\cite{Aydemir2025ARXIV} & 61.7 & \underline{89.9} & 68.1 & 80.5 & \underline{93.4} & 55.3 & 69.3 & \underline{89.6} & \underline{67.0} & \underline{79.9} & \underline{92.0} \\ 
        \midrule

        \textbf{Real-World Fine-Tuning} \\
        BootsTAPIR~\cite{Doersch2024ARXIV} & 55.7 & 78.2 & 64.9 & 80.1 & 86.3 & 54.6 & 68.4 & 86.5 & 61.4 & 73.6 & 88.7 \\
        Anthro-LocoTrack~\cite{Kim2025ARXIV} & 61.1 & 89.5 & 64.7 & 79.2 & 88.4 & 53.9 & 68.4 & 86.4 & 64.8 & 77.3 & 89.1 \\ 
        BootsTAPNext-B~\cite{Zholus2025ICCV} & 33.6 & 68.5 & 64.0 & 75.0 & 88.7 & \underline{57.3} & \underline{70.6} & 87.4 & 65.2 & 78.5 & 91.2 \\
        CoTracker3 (Window)~\cite{Karaev2024ARXIV} & 54.0 & 84.4 & 66.4 & 78.8 & 90.8 & 55.8 & 68.5 & 88.3 & 63.8 & 76.3 & 90.2 \\
        AllTracker$^\dagger$~\cite{Harley2025ICCV} & \underline{62.0} & 87.1 & \underline{68.8} & \underline{80.9} & 92.2 & 56.8 & 69.3 & 89.1 & 63.7 & 77.0 & 88.7 \\
        Track-On-R (Ours) & \textbf{67.3} & \textbf{90.2} & \textbf{70.9} & \textbf{82.6} & \textbf{94.0} & \textbf{57.8} & \textbf{71.0} & \textbf{90.5} & \textbf{68.1} & \textbf{80.3} & \textbf{92.5} \\
        \bottomrule
    \end{tabular}
\end{table*}

\subsection{Setup}
\label{sec:exp:experimental_setup}

\vspace{3pt} \noindent\textbf{Datasets.}  
For verifier training, we use K-EPIC~\cite{Darkhalil2025WACV}, a synthetic dataset of 11K videos, each 24 frames long, containing similar synthetic objects to those in TAP-Vid Kubric~\cite{Doersch2022NeurIPS}.  
For real-world fine-tuning with the verifier, we use 8K videos from our filtered collection described in~\secref{sec:verifier:fine_tuning}.  
We evaluate our model on four public real-world datasets with diverse characteristics: 
{TAP-Vid DAVIS}, 30 real-world videos from the DAVIS dataset;  
{TAP-Vid Kinetics}, 1000 videos from the validation split of Kinetics-700-2020;  
{RoboTAP}~\cite{Vecerik2023ICRA}, 265 robotic sequences averaging over 250 frames each; and  
{EgoPoints}~\cite{Darkhalil2025WACV}, long egocentric videos spanning up to several thousand frames.

\vspace{3pt} \noindent\textbf{Metrics.}  
We evaluate tracking performance on TAP-Vid subsets and RoboTAP using three standard metrics:  
{Occlusion Accuracy (OA)}, visibility prediction accuracy; \deltaavg, the average proportion of visible points tracked within thresholds of 1, 2, 4, 8, and 16 pixels; and  
{Average Jaccard (AJ)}, a combined measure of localization and visibility.  
On EgoPoints, we report $\delta_{\text{avg}}$ and OA following the official evaluation.

\vspace{3pt} \noindent\textbf{Evaluation details.}  
We follow the standard TAP-Vid benchmarking protocol, downsampling all videos to $256 \times 256$ resolution except for EgoPoints.  
Models are evaluated in the {queried-first} setting, corresponding to the causal tracking scenario:  
the first visible point in each trajectory is used as the query, and the model tracks that point in subsequent frames.  
For EgoPoints, frames are resized to $384 \times 512$, however evaluation is done on $256 \times 256$, consistent with its official evaluation setup.

\subsection{Results: Verifier as Inference-time Ensemble}
\label{sec:exp:verifier}

We first evaluate the verifier as an inference-time ensemble to assess its ability to identify the most reliable tracker per frame. 
As shown in~\figref{fig:verifier_comparison}, we compare the \textbf{reproduced} performance of each teacher model using their official checkpoints, 
the random-teacher baseline (where a teacher is randomly selected per video), and the verifier-based selection across four real-world datasets. 
Reproduced results are used for all comparisons. These correspond to the exact candidate trajectories provided to the verifier, ensuring that frame-wise selection is evaluated fairly.

The verifier consistently outperforms both the random baseline and the strongest individual teacher.
This demonstrates its ability to adaptively select the most reliable prediction at each time step. 
The gains are particularly pronounced on challenging datasets such as EgoPoints, where model reliability fluctuates significantly across motion patterns and scene content.

Performance rankings also vary substantially across datasets. 
For example, BootsTAPNext~\cite{Zholus2025ICCV} ranks last on RoboTAP but second on DAVIS, highlighting that no single tracker is universally dominant. 
These results confirm that the verifier provides a principled mechanism for aggregating complementary teacher models and serves as a robust adaptive ensemble strategy, validating its design prior to real-world fine-tuning.

\subsection{Results: Verifier-Guided Adaptation}
\label{sec:exp:real_world}

In this section, we compare our model \textbf{Track-On-R}, fine-tuned on real-world videos using the verifier-guided approach, against prior work.
As summarized in~\tabref{tab:sota_A}, we categorize existing approaches into two groups: \textbf{synthetic pretraining} and \textbf{real-world fine-tuning}, presented side by side for direct comparison.
Models in the first group are trained exclusively on synthetic data from TAP-Vid Kubric~\cite{Doersch2022NeurIPS},
whereas those in the second group leverage additional real-world videos for fine-tuning to enhance generalization.
When official results are available, we report the numbers published by the original authors.
If a method has no officially reported values, we reproduce the results using the authors’ released checkpoints.
For methods without any publicly released checkpoints or reported values, we indicate missing entries with “—”.

\vspace{3pt}
\noindent\textbf{Baseline models.}
Existing real-world adaptation strategies differ in both supervision and data scale. BootsTAPIR~\cite{Doersch2024ARXIV} and BootsTAPNext~\cite{Zholus2025ICCV} rely on large-scale student–teacher distillation over millions of real videos. AnthroTAP~\cite{Kim2025ARXIV} instead uses domain-specific human-mesh annotations to derive pseudo-labels, which limits its applicability outside human-centric datasets. AllTracker~\cite{Harley2025ICCV} does not perform real-world fine-tuning, but is trained on high-quality synthetic optical flow with ground-truth as well as annotated real-world flow data. The most similar approach is CoTracker3~\cite{Karaev2024ARXIV}, which applies a generic random-teacher pseudo-labeling strategy on real-world videos. Our method also follows a teacher-based pseudo-labeling paradigm, but introduces a verifier that enables reliable adaptation to arbitrary real-world video collections using only raw videos.

\vspace{3pt} \noindent\textbf{Comparison on EgoPoints.}
EgoPoints presents a particularly challenging setting due to long sequences and extended occlusion periods.
Our model significantly improves over the synthetic baseline, increasing \deltaavg from 61.7 to 67.3.
It also widens the gap to the closest competitor, AllTracker~\cite{Harley2025ICCV}, by +5.3 in \deltaavg.
These results demonstrate that our pseudo-labels transfer effectively to complex ego-centric motion, even though the training data are not explicitly ego-centric.

\vspace{3pt}
\noindent\textbf{Comparison on RoboTAP.}
Compared to the synthetic baseline, our model achieves consistent gains across all metrics, including a +2.8 improvement in AJ.
Our adaptation attains the highest scores on the benchmark, outperforming both synthetic and real-world baselines.
Notably, our real-world training set contains no robotic sequences, indicating that the verifier-guided pseudo-labels generalize well and transfer effectively to robotic manipulation scenarios.

\vspace{3pt}
\noindent\textbf{Comparison on Kinetics.}
On TAP-Vid Kinetics, our model improves over the synthetic baseline by +2.5 in AJ. It also surpasses the highest AJ score of 57.3 reported by BootsTAPNext~\cite{Zholus2025ICCV}, while achieving 3.1 points higher in OA, indicating more accurate visibility prediction.

\vspace{3pt}
\noindent\textbf{Comparison on DAVIS.}
Our model achieves the highest scores across all metrics, outperforming both synthetic and real-world baselines.
Compared to the baseline Track-On2~\cite{Aydemir2025ARXIV}, our fine-tuned variant improves AJ by +1.1, establishing a new state of the art on this benchmark.

\subsection{Ablation Study}
\label{sec:exp:ablation}

\begin{table}
    \centering
    \small
    \caption{\textbf{Effect of teacher composition on verifier performance.}
    We report \deltaavg for the random-teacher baseline (Rand.) and the verifier-based selection (Ver.) across different teacher subsets.
    }
    \label{tab:verifier_ensemble}
    \begin{tabular}{l ccccc cc cc}
        \toprule
        
        & \multicolumn{5}{c}{\textbf{Teacher Models}} & 
        \multicolumn{2}{c}{DAVIS}  &
        \multicolumn{2}{c}{RoboTAP} \\
        \cmidrule(r){2-6}  \cmidrule(r){7-8}  \cmidrule(r){9-10}  
        & A & B & C & D & E & Rand. & Ver. & Rand. & Ver.
        \\
        \midrule
        (i) & \cmark & \cmark & \xmark & \xmark & \xmark & 79.5 & 80.6 & 77.4 & 81.8 \\
        (ii) & \cmark & \cmark & \cmark & \xmark & \xmark & 79.4 & 80.6 & 77.3 & 82.4 \\
        (iii) & \cmark & \cmark & \cmark & \cmark & \xmark & 77.5 & 80.8 & 77.4 & 82.8 \\ 
        (iv) & \cmark & \cmark & \cmark & \cmark & \cmark & 77.7 & 81.1 & 78.0 & 83.1 \\
       \bottomrule
    \end{tabular}
\end{table}

\vspace{3pt} \noindent\textbf{Teacher models.}
To analyze how the composition of teacher models influences performance, \tabref{tab:verifier_ensemble} compares the random selection (round-robin selection) and our verifier-based selection over different teacher subsets in terms of \deltaavg on TAP-Vid DAVIS and RoboTAP. 
Models A–E correspond to Track-On2~\cite{Aydemir2025ARXIV}, BootsTAPNext~\cite{Zholus2025ICCV}, BootsTAPIR~\cite{Doersch2024ARXIV}, CoTracker~\cite{Karaev2024ARXIV}, and Anthro-LocoTrack~\cite{Kim2025ARXIV}.
The verifier consistently outperforms the random baseline and remains robust even when weaker teachers are included. 
Adding or removing a model that lowers the baseline average either preserves or improves verifier accuracy.
For example, in DAVIS, adding D to \{A,B,C\} (rows ii–iii) lowers the random baseline but increases verifier accuracy. 
These trends indicate that the verifier effectively exploits complementary teacher behaviors rather than being diluted by weaker models.

\vspace{3pt} \noindent \textbf{Synthetic–real training schedule.}
We study the effect of training data composition during fine-tuning in~\tabref{tab:syn_real_training}. We compare three settings: using only real-world videos from $\cD_{\text{real}}$, mixing $\cD_{\text{real}}$ with synthetic data $\cD_{\text{syn}}$ (with ground-truth supervision), and the same mixture with a schedule that gradually increases the loss weight of real-world samples (our default). The mixture improves OA due to visibility supervision from synthetic data, while real-only training yields slightly better localization in terms of \deltaavg. The scheduled mixture combines both advantages and achieves the best overall performance, although adding synthetic data on top of real-only training provides only marginal gains.

\begin{table}
    \centering
    \small
    \setlength{\tabcolsep}{3pt}
    \caption{\textbf{Synthetic \vs real data during fine-tuning.}
    We compare three configurations: Real (only $\cD_{\text{real}}$), Mix ($\cD_{\text{real}}+\cD_{\text{syn}}$), and Mix + Schedule, where the loss weight of real-world samples is gradually increased during training.
    }
    \label{tab:syn_real_training}
    \begin{tabular}{l cc ccc ccc ccc}
        \toprule
        \multirow{3}{*}{\textbf{Data}} 
        & \multicolumn{2}{c}{EgoPoints} 
        & \multicolumn{2}{c}{RoboTAP} 
        & \multicolumn{2}{c}{Kinetics} 
        & \multicolumn{2}{c}{DAVIS} \\
        \cmidrule(r){2-3} \cmidrule(r){4-5} \cmidrule(r){6-7} \cmidrule(r){8-9}
         & \deltaavg & OA
         & \deltaavg & OA 
         & \deltaavg & OA 
         & \deltaavg & OA  \\
        \midrule
        Real        & 66.9 & 90.2 & \textbf{82.7} & 93.8 & 70.9 & 90.4 & 80.3 & 92.4 \\
        Mix         & 65.6 & \textbf{90.7} & 82.3 & 93.9 & 70.8 & 90.3 & \textbf{80.7} & \textbf{92.5} \\
        Mix + Schedule  & \textbf{67.3} & 90.2 & 82.6 & \textbf{94.0} & \textbf{71.0} & \textbf{90.5} & 80.3 & \textbf{92.5} \\
        \bottomrule
    \end{tabular}
\end{table}

\section{Conclusion \& Limitation}
\label{sec:discussion}

In this paper, we have introduced a verifier module that learns to assess the reliability of point tracker predictions, enabling robust real-world adaptation without manual annotations.
When used for fine-tuning, it produces high-quality pseudo-labels that substantially improve tracking performance across diverse domains and achieve data-efficient adaptation.
Beyond pseudo-label generation, the verifier also functions as a effective inference-time ensemble mechanism, offering a general framework for model selection and uncertainty estimation in video correspondence tasks.
However, the effectiveness of fine-tuning still depends on the quality and diversity of the available video data, which highlights the importance of curated real-world collections.
In addition, the verifier’s upper bound is constrained by the quality of its teacher trackers, suggesting that future work should focus on developing stronger and more robust pretrained models to further enhance verifier-guided learning.

\section*{Acknowledgement}
This project is funded by the European Union (ERC, ENSURE, 101116486) with additional compute support from Leonardo Booster (EuroHPC Joint Undertaking, EHPC-AI-2024A05-028). Views and opinions expressed are however those of the author(s) only and do not necessarily reflect those of the European Union or the European Research Council. Neither the European Union nor the granting authority can be held responsible for them. Weidi would like to acknowledge the funding from Scientific Research Innovation Capability Support Project for Young Faculty~(ZY-GXQNJSKYCXNLZCXM-I22).

{
    \small
    \bibliographystyle{ieeenat_fullname}
    \bibliography{bibliography_long, main}
}

\clearpage
\setcounter{page}{1}
\maketitlesupplementary

In this supplementary material, we provide additional implementation details, analyses, and visualizations that complement the main paper. 
\secref{appendix:verifier_details} introduces the verifier model in full, including its architecture, training setup, and trajectory perturbation strategy. 
\secref{appendix:real_world_fine_tuning} details the real-world adaptation pipeline and fine-tuning procedure. 
\secref{appendix:additional_ablations} presents additional ablation studies. 
Finally, qualitative examples and visualizations of the verifier are provided in~\secref{sec:visualizations}.

\section{Verifier Details}  
\label{appendix:verifier_details}

This section provides the complete specification of the verifier. We describe its architecture, training procedure, and the perturbation strategy used to synthesize diverse candidate trajectories during training.

\subsection{Training Setup}
\label{appendix:verifier_details:training}

\noindent\textbf{Training schedule.}
The verifier is trained for 100 epochs on K-EPIC~\cite{Darkhalil2025WACV}, corresponding to approximately 36K iterations with batch size 32. Training uses full 24-frame clips from K-EPIC, each providing up to 384 tracks.

\vspace{3pt}
\noindent\textbf{Optimization.}
Optimization uses AdamW~\cite{Loshchilov2019ICLR} on 32$\times$A100 (64\,GB) GPUs with mixed-precision training. 
The learning rate follows a cosine decay schedule with 1\% warmup and peaks at $5{\times}10^{-4}$. 
Weight decay is set to $1{\times}10^{-3}$, and gradient norms are clipped to 1.0.

\vspace{3pt}
\noindent\textbf{Data augmentation.}
We apply random resized cropping with scale range $[0.6,1.0]$, horizontal flipping, and color jittering in brightness, contrast, saturation, and hue. Additional augmentations include Gaussian blur, low-probability solarization, and JPEG compression with a randomly sampled quality factor. Small geometric perturbations such as rotations up to $10^\circ$ and mild perspective distortions are also applied. All augmentations are applied consistently across frames of a clip.

\subsection{Architecture}
\label{appendix:verifier_details:architecture}

\begin{figure}[t]
    \centering
    \includegraphics[width=1\linewidth]{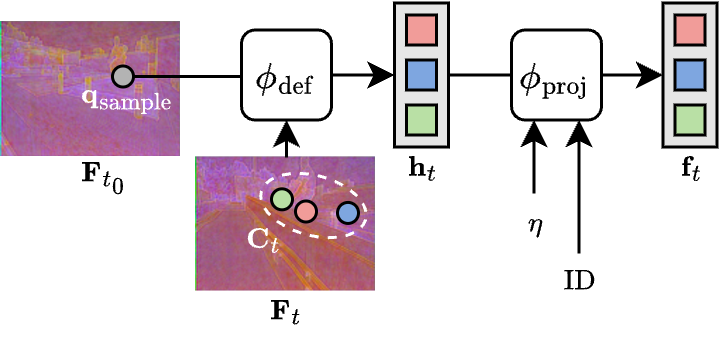}
    \caption{\textbf{Localized Feature Extraction.}
    Given frame-wise features of the query frame $t_0$ and target frame $t$, denoted by $\bF_{t_0}$ and $\bF_t$, we first bilinearly sample the reference feature $\bq_{\text{sample}}$ at the query location. A deformable attention module $\phi_{\text{def}}$ then aggregates localized context around each candidate location ($\bC_t$), producing descriptors $\bh_t$. We concatenate displacement embeddings $\eta(\cdot)$ with an identity embedding (query vs.\ candidate) and project via $\phi_{\text{proj}}$ to obtain the final query and candidate features consumed by the candidate transformer.}
    \label{fig:localized_feature_extraction}
\end{figure}

\noindent \textbf{Visual encoder.} A stride-4 CNN encoder is initialized from the CoTracker3 video variant~\cite{Karaev2024ARXIV} and projected to the model width $D{=}256$ using a $1{\times}1$ convolution. A four-level feature hierarchy is constructed by average pooling the stride-4 feature map with factors ${2,4,8}$, producing features at strides ${4,8,16,32}$. Feature sampling (as defined in the main paper) is performed at the highest resolution. These multi-scale features provide the input to a three-layer deformable attention decoder that extracts localized descriptors around query and candidate positions.

\vspace{3pt} \noindent \textbf{Localized feature extraction.}
As illustrated in~\figref{fig:localized_feature_extraction}, we first bilinearly sample a reference feature from the query-frame feature map $\bF_{t_0}$ at the query location. Conditioned on this reference, a three-layer deformable attention decoder extracts localized descriptors by attending to regions centered at the query (at $t_0$) and at all candidate locations (at frame $t$) over the corresponding feature maps. Spatial displacements are encoded using a 2D sinusoidal embedding $\eta(\cdot)$ of size 32, concatenated with raw $(x,y)$ coordinates and scaled by a learnable factor initialized to 16. A binary identity embedding distinguishes query tokens from candidate tokens. The resulting representations are concatenated, normalized, and projected to the model width. A learnable temporal embedding, initialized for 24-frame clips, is interpolated when clip lengths vary.

\vspace{3pt} \noindent \textbf{Candidate transformer.} Each multi-head attention block uses four attention heads followed by residual connections and normalization. The feed-forward network expands the hidden dimension by a factor of four, applies a GELU activation, and uses a dropout rate of 0.1 before a final normalization step. The module follows a post-normalization transformer design.

\vspace{3pt} \noindent \textbf{Verifier output.} A ranking head linearly projects both query and candidate features, applies L2 normalization, and computes their dot-product similarity. A learnable temperature parameter, initialized to $0.1$, scales the resulting logits to form the frame-wise reliability distribution. \\

\subsection{Track Augmentations}
\label{appendix:verifier_details:track_augmentations}

To approximate realistic tracker behavior, we perturb ground-truth trajectories with stochastic transformations to generate candidate predictions during training. Each trajectory starts with light one-pixel Gaussian noise to model localization uncertainty. A random subset of the following perturbations is then applied, with each triggered by a fixed probability $p$:

\begin{enumerate}
\item \textbf{Stable noise ($p=0.5$):} Small Gaussian displacements (2–4 pixels) smoothed across 3–5 frames.
\item \textbf{Gradual drift ($p=0.4$):} Progressive displacement or soft identity blending with nearby visible points (within 16–32 pixels).
\item \textbf{Long-term drift ($p=0.3$):} A steadily increasing offset up to 64 pixels.
\item \textbf{Spiky noise ($p=0.3$):} Short, high-magnitude spikes of about 8 pixels that recover quickly.
\item \textbf{Abrupt jump ($p=0.1$):} Sudden jumps or identity switches up to 128 pixels.
\item \textbf{Complete switch ($p=0.1$):} Replacement of an entire trajectory with another visible one.
\end{enumerate}

Multiple perturbations may co-occur, producing diverse deviations ranging from 1 to 128 pixels. This strategy exposes the verifier to the full range of errors observed in real trackers and encourages robust reliability estimation.

\section{Real-world Fine-tuning Pipeline}  
\label{appendix:real_world_fine_tuning}

This section describes the full configuration of our real-world fine-tuning pipeline, including training details and the teacher model configurations.

\subsection{Training Setup}
\label{appendix:real_world_fine_tuning:training}

\noindent\textbf{Training schedule.}
We fine-tune the pretrained Track-On2~\cite{Aydemir2025ARXIV} model for 24 epochs (approximately 3.7K iterations) with batch size 32 using a mixture of TAP-Vid Kubric~\cite{Karaev2024ARXIV} and the collected real-world dataset described in~\secref{sec:verifier:fine_tuning}. 
During training, both datasets initially contribute equally to the loss. 
The loss weight of real-world videos is then linearly increased from 1.0 to 2.0, while the synthetic weight is reduced to 0. 
This schedule gradually shifts supervision toward real-world data.

\vspace{3pt}
\noindent\textbf{Supervision.}
For synthetic videos, ground-truth trajectories and visibility annotations are available, and the model is supervised using all training losses. 
For real-world videos, we instead use verifier-guided pseudo-labels. 
Occluded frames are masked for localization losses based on majority-voted visibility across teacher predictions. 
The visibility and uncertainty heads are not explicitly supervised for real-world data. 
All remaining hyperparameters, including memory size and top-$k$ candidate selection, follow the original Track-On2 training configuration.

\vspace{3pt}
\noindent\textbf{Optimization.}
Training uses the AdamW optimizer~\cite{Loshchilov2019ICLR} on 32 A100 GPUs (64\,GB) with mixed-precision training. 
The learning rate follows a cosine decay schedule without warmup, with a peak value of $3{\times}10^{-5}$. 
Weight decay is set to $1{\times}10^{-5}$, and gradient norms are clipped to 1.0.

\vspace{3pt}
\noindent\textbf{Query sampling and data preparation.}
Training uses clips of length 48 with 256 query points per video. 
For real-world videos, two thirds of the queries are sampled from SIFT~\cite{Lowe1999ICCV} detections, while the remaining queries are selected from motion-salient regions obtained via grayscale frame differencing with mild spatial smoothing. 
This lightweight detector reliably identifies moving objects and reduces the proportion of static or low-texture points. 
Queries are extracted from the first 24 frames with temporal stride 4 (frames 1, 5, $\dots$, 25). 
If fewer than 256 keypoints are detected, the remaining locations are filled with randomly sampled points. 
We apply the same video-level augmentations used for verifier training (see~\secref{appendix:verifier_details:training}) consistently across all frames of a clip. 
Teacher predictions are computed on the clean video (without augmentation) to obtain pseudo-label supervision.

\subsection{Teacher Models and Reproduction}
\label{appendix:baseline_models}

We evaluate all teacher methods using the official checkpoints released by their respective authors. Unless stated otherwise, each model is executed with its recommended inference configuration. This ensures that our reproduced results reflect the intended operating conditions of each method and provide a fair comparison. Specific implementation details are summarized below:

\vspace{3pt} \noindent \textbf{Track-On2}~\cite{Aydemir2025ARXIV}. We use the official DINOv3~\cite{Simeoni2025arXiv}-based checkpoint with memory size 24. A global grid of size 10 is employed when generating pseudo-labels.

\vspace{3pt} \noindent \textbf{BootsTAPIR}~\cite{Doersch2024ARXIV} \& \textbf{BootsTAPNext-B}~\cite{Zholus2025ICCV}. Evaluated directly using the released checkpoints in \texttt{PyTorch}~\cite{Paszke2019NeurIPS} with their default inference settings.

\vspace{3pt} \noindent \textbf{CoTracker3}~\cite{Karaev2024ECCV}. The publicly available checkpoints correspond to real-world fine-tuned models. We use the window-input variant with a global grid of size 10 and joint multi-point tracking enabled.

\vspace{3pt} \noindent \textbf{Anthro-LocoTrack}~\cite{Cho2024ECCV}. Base model evaluated at $256{\times}256$ resolution.

\vspace{3pt} \noindent \textbf{AllTracker}~\cite{Harley2025ICCV}. Model trained on Kubric + Mix, evaluated at $384{\times}512$ resolution. As it is not a sparse point tracker, we run the model independently for each frame containing query points.

\section{Additional Experiments}
\label{appendix:additional_ablations}

\begin{table}[t]
    \centering
    \small
    \setlength{\tabcolsep}{5pt}
    \caption{\textbf{Quantitative results on synthetic benchmarks.} 
    We compare the synthetic baseline Track-On2 with our real-world fine-tuned variant (Track-On-R) on Dynamic Replica (DR) and PointOdyssey.}
    \label{tab:sota_syn}
    \begin{tabular}{l c ccc}
        \toprule
        \multirow{3}{*}{\textbf{Model}} 
        & \multicolumn{1}{c}{DR} 
        & \multicolumn{3}{c}{PointOdyssey} \\
        \cmidrule(r){2-2} \cmidrule(r){3-5}
        & \deltaavg \up 
        & \deltaavg \up & MTE $\downarrow$ & Survival \up \\
        \midrule
        BootsTAPNext & 46.2 & 9.9 & 88.2 & 12.8 \\
        CoTracker3 & 72.3 & 44.5 & 20.7 & 56.3 \\
        Track-On2 & 74.5 & 45.1 & 22.0 & 57.7 \\
        Track-On-R (Ours) & \textbf{75.1} & \textbf{53.4} & \textbf{13.4} & \textbf{63.1} \\
        \bottomrule
    \end{tabular}
\end{table}

\begin{table}[b]
    \centering
    \caption{\textbf{Non-learning ensemble baselines \vs verifier.}
    Comparison of fixed ensemble heuristics and the learned verifier on four benchmarks measured by \deltaavg.}
    \setlength{\tabcolsep}{3pt}
    \label{tab:rebuttal_nonlearning_baselines}
    \small
    \begin{tabular}{l cccc}
         \toprule
         \textbf{Ensemble} & EgoPoints & RoboTAP & Kinetics & DAVIS \\
         \midrule
         Geometric median     & 60.1 & 81.7 & 70.3 & 80.6  \\
         Agreement-based      & 61.8 & 81.8 & 70.4 & 80.5 \\
         Kalman constant-vel. & 52.1 & 76.5 & 65.8 & 72.8 \\
         Minimum acceleration & 54.0 & 78.3 & 67.2 & 77.8 \\
         \midrule
         Verifier & 64.8 & 82.8 & 71.2 & 80.8 \\
       \bottomrule
    \end{tabular}
\end{table}

\noindent \textbf{Comparison on synthetic benchmarks.}
In~\tabref{tab:sota_syn}, we compare our real-world fine-tuned model with the previous work on the synthetic benchmarks Dynamic Replica~\cite{Karaev2023CVPR} and PointOdyssey~\cite{Zheng2023ICCV}. 
Dynamic Replica consists of 300-frame videos, while PointOdyssey contains extremely long sequences with thousands of frames, making it particularly suitable for evaluating long-term temporal robustness. 
Although our model is fine-tuned using only real-world videos, it still improves performance on synthetic benchmarks. 
On Dynamic Replica, the gain is modest (+0.6 \deltaavg). 
In contrast, the improvement on PointOdyssey is substantial, with +8.3 in \deltaavg and +5.4\% in survival rate. 
These results indicate that our real-world adaptation does not degrade synthetic-domain performance, and in fact improves long-term tracking robustness, particularly on very long sequences.

\vspace{3pt}
\noindent \textbf{Non-learning baselines.}
In~\tabref{tab:rebuttal_nonlearning_baselines}, we compare the verifier against representative non-learning ensemble heuristics: (i) geometric median (Weiszfeld’s algorithm); (ii) agreement-based selection, which selects the candidate with minimum average distance to others; (iii) a Kalman-style constant-velocity selector favoring temporal smoothness; and (iv) a minimum-acceleration selector that chooses the candidate closest to constant-velocity extrapolation from the previous two frames.
We use four teacher models in the ensemble (Track-On2, BootsTAPNext, BootsTAPIR, and CoTracker3), whose predictions often disagree on challenging frames, making selection non-trivial. While spatial consensus methods outperform temporal heuristics, the learned verifier consistently surpasses all fixed strategies, with the largest margins observed in challenging settings such as EgoPoints. This indicates that rigid heuristics fail to capture diverse failure modes, whereas the verifier adapts to varying error patterns.

\vspace{3pt} 
\noindent \textbf{The $\cD_{\text{real}}$ collection.}
To analyze the impact of the real-world dataset $\cD_{\text{real}}$ used for fine-tuning, we compare models trained on TAO only (2921 videos) and on the full collection of TAO, OVIS, and VSPW described in~\secref{sec:verifier:fine_tuning} (4864 videos), as summarized in~\tabref{tab:d_real_content}. 
For this experiment, we use only real-world data without additional synthetic supervision. 
We observe that adding OVIS and VSPW on top of TAO yields only marginal improvements, with most gains below 0.2 points. 
In some cases, the TAO-only model performs slightly better (\eg, OA on EgoPoints). 
These results suggest that most of the adaptation benefit can already be achieved with fewer than 3K real videos and short fine-tuning, highlighting the efficiency of our approach for domain adaptation.

\begin{table}[t]
    \centering
    \small
    \setlength{\tabcolsep}{3.5pt}
    \caption{\textbf{Effect of the $\cD_{\text{real}}$ dataset composition.}
    We compare fine-tuning using TAO only (2921 videos) versus the full collection of TAO, OVIS, and VSPW (4864 videos). }
    \label{tab:d_real_content}
    \begin{tabular}{l cc ccc ccc ccc}
        \toprule
        \multirow{3}{*}{$\cD_{\text{real}}$} 
        & \multirow{3}{*}{\textbf{Size}} 
        & \multicolumn{2}{c}{EgoPoints} 
        & \multicolumn{2}{c}{RoboTAP} 
        & \multicolumn{2}{c}{Kinetics} 
        & \multicolumn{2}{c}{DAVIS} \\
        \cmidrule(r){3-4} \cmidrule(r){5-6} \cmidrule(r){7-8} \cmidrule(r){9-10}
         & 
         & \deltaavg & OA
         & \deltaavg & OA 
         & \deltaavg & OA 
         & \deltaavg & OA  \\
        \midrule
        TAO & 2.9K & 66.1 & \textbf{91.1} & 82.6 & 93.6 & 70.8 & 90.2 & \textbf{80.3} & 92.3 \\
        Mix & 4.9K & \textbf{66.9} & 90.2 & \textbf{82.7} & \textbf{93.8} & \textbf{70.9} & \textbf{90.4} & \textbf{80.3} & \textbf{92.4} \\
        \bottomrule
    \end{tabular}
\end{table}

\section{Visualizations}
\label{sec:visualizations}

\begin{figure*}[h]
  \centering
  \begin{subfigure}[t]{\linewidth}
    \centering
    \includegraphics[width=\linewidth]{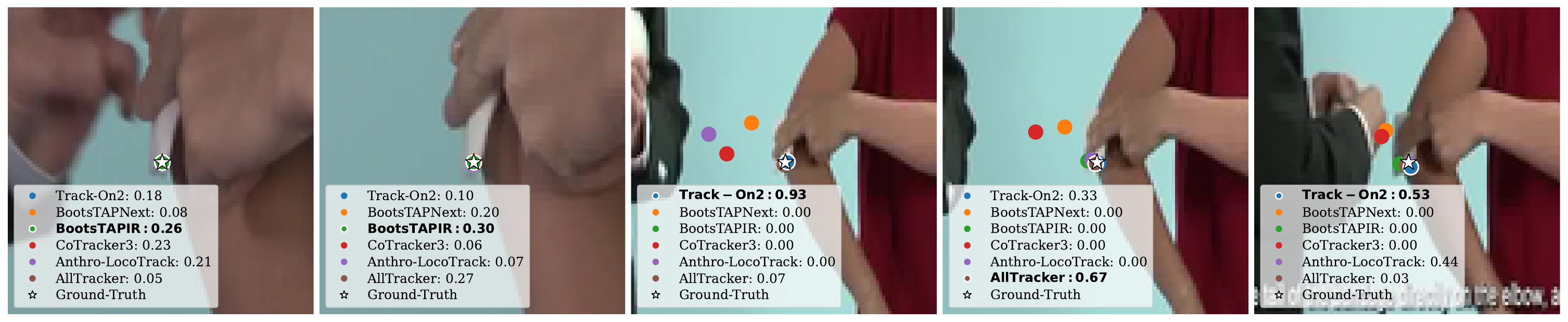}
  \end{subfigure}
  \begin{subfigure}[t]{\linewidth}
    \centering
    \includegraphics[width=\linewidth]{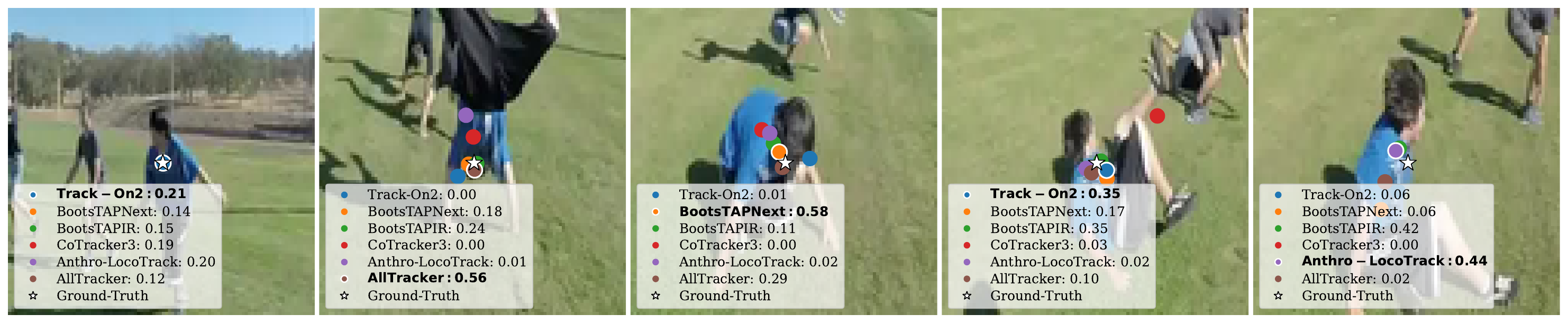}
  \end{subfigure}
  \begin{subfigure}[t]{\linewidth}
    \centering
    \includegraphics[width=\linewidth]{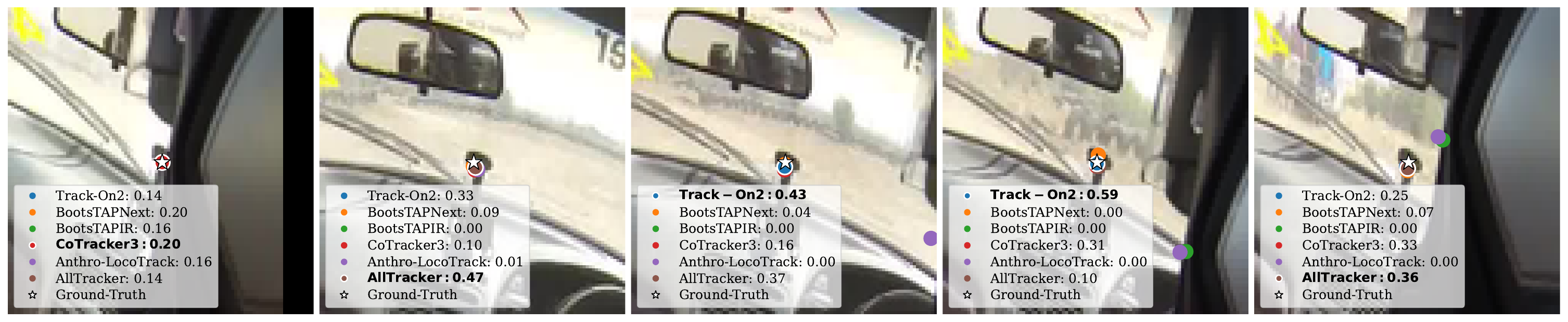}
  \end{subfigure}
    \caption{\textbf{Verifier selection behavior across videos.}
    Each row corresponds to a different video from TAP-Vid Kinetics. 
    Frames are uniformly sampled among visible ones, and a $50\times50$ crop centered at the ground-truth point is shown. 
    Colored dots indicate predictions from teacher trackers, the star marks the ground-truth location, and the legend lists the verifier reliability scores, with the selected candidate highlighted in bold. 
    The verifier adaptively switches between trackers across frames, assigning higher scores to spatially accurate predictions while suppressing unreliable ones.}
  \label{fig:ensemble_examples}
\end{figure*}

We visualize examples of the verifier’s per-frame selection behavior on videos from the TAP-Vid Kinetics dataset in~\figref{fig:ensemble_examples}. Each row corresponds to a different video and illustrates how the verifier dynamically assigns reliability scores to candidate predictions over time. For visualization, we uniformly sample frames among the visible ones and show a $50 \times 50$ crop centered at the ground-truth point. Predictions from teacher models are shown as colored dots, the ground-truth location is marked with a star, and the verifier scores are listed in the legend, where the selected candidate is highlighted in bold.

First, we observe that the verifier selects different trackers across frames rather than consistently following a single globally strong model, which aligns with its intended design. Second, the verifier assigns higher scores to spatially accurate predictions while suppressing incorrect ones, \eg \textcolor{darkorange}{BootsTAPNext}, \textcolor{red}{CoTracker3}, and \textcolor{mediumpurple}{Anthro-LocoTrack} in the first example (top row).

Even when the selected prediction is not strictly the best among the available candidates, the verifier tends to identify the group of reliable predictions and selects among them. For instance, in the second example (middle row), the verifier chooses between \textcolor{darkorange}{BootsTAPNext}, \textcolor{forestgreen}{BootsTAPIR}, and \textcolor{darkbrown}{AllTracker}, which all remain close to the ground-truth. Although the exact ranking between these models may differ by a few pixels, the verifier consistently favors this reliable group while avoiding large tracking errors.

\end{document}